\documentclass[runningheads]{llncs}

\usepackage{graphicx}
\usepackage{amsmath}
\usepackage{amssymb}
\usepackage[hyphens]{url}

\begin{document}

\title{Heavy-tailed kernels reveal a finer cluster structure in t-SNE visualisations}

\author{Dmitry Kobak\inst{1} \and
George Linderman\inst{2} \and
Stefan Steinerberger\inst{3} \and
Yuval Kluger\inst{2,4} \and
Philipp Berens\inst{1}}

\authorrunning{D. Kobak et al.}

\institute{Institute for Ophthalmic Research, University of T\"{u}bingen, Germany \\
\email{\{dmitry.kobak, philipp.berens\}@uni-tuebingen.de} \and
Applied Mathematics Program, Yale University, New Haven, USA \and
Department of Mathematics, Yale University, New Haven, USA \and
Department of Pathology, Yale School of Medicine, New Haven, USA \\
\email{\{george.linderman, stefan.steinerberger, yuval.kluger\}@yale.edu}}

\maketitle

\begin{abstract}
T-distributed stochastic neighbour embedding (t-SNE) is a widely used data visualisation technique. It differs from its predecessor SNE by the low-dimensional similarity kernel: the Gaussian kernel was replaced by the heavy-tailed Cauchy kernel, solving the ``crowding problem'' of SNE. Here, we develop an efficient implementation of t-SNE for a $t$-distribution kernel with an arbitrary degree of freedom $\nu$, with $\nu\to\infty$ corresponding to SNE and $\nu=1$ corresponding to the standard t-SNE. Using theoretical analysis and toy examples, we show that $\nu<1$ can further reduce the crowding problem and reveal finer cluster structure that is invisible in standard t-SNE. We further demonstrate the striking effect of heavier-tailed kernels on large real-life data sets such as MNIST, single-cell RNA-sequencing data, and the HathiTrust library. We use domain knowledge to confirm that the revealed clusters are meaningful. Overall, we argue that modifying the tail heaviness of the t-SNE kernel can yield additional insight into the cluster structure of the data.

\end{abstract}

\section{Introduction}

T-distributed stochastic neighbour embedding (t-SNE) \cite{maaten2008visualizing} and related methods \cite{tang2016visualizing,mcinnes2018umap} are used for data visualisation in many scientific fields dealing with thousands or even millions of high-dimensional samples. They range from single-cell cytometry \cite{amir2013visne} and transcriptomics \cite{tasic2018shared,zeisel2018molecular}, where samples are cells and features are proteins or genes, to population genetics \cite{diaz2018revealing}, where samples are people and features are single-nucleotide polymorphisms, to humanities \cite{schmidt2018stable}, where samples are books and features are words.

T-SNE was developed from an earlier method called SNE \cite{hinton2003stochastic}. The central idea of SNE was to describe pairwise relationships between high-dimensional points in terms of normalised affinities: close neighbours have high affinity whereas distant samples have near-zero affinity. SNE then positions the points in two dimensions such that the Kullback-Leibler divergence between the high- and low-dimensional affinities is minimised. This worked to some degree but suffered from what was later called the ``crowding problem''. The idea of t-SNE was to adjust the kernel  transforming pairwise low-dimensional distances into affinities: the Gaussian kernel was replaced by the heavy-tailed Cauchy kernel (t-distribution with one degree of freedom $\nu$), ameliorating the crowding problem.

The choice of the specific heavy-tailed kernel was mostly motivated by mathematical and computational simplicity: a t-distribution with $\nu=1$ has a density proportional to $1/(1+x^2)$ which is mathematically compact and fast to compute. However, a t-distribution with any finite $\nu$ has heavier tails than the Gaussian distribution (which corresponds to $\nu\to\infty$). It is therefore reasonable to explore the whole spectrum of the values of $\nu$ from $\infty$ to 0. Given that t-SNE ($\nu=1$) outperforms SNE ($\nu=\infty$), it might be that for some data sets $\nu<1$ would offer additional insights into the structure of the data. 

While this seems like a straightforward extension and has already been discussed in the literature \cite{maaten2009learning,yang2009heavy}, no efficient implementation of this idea has been available until now. T-SNE is usually optimised via adaptive gradient descent. While it is easy to write down the gradient for an arbitrary value of $\nu$, the exact t-SNE from the original paper requires $\mathcal O(n^2)$ time and memory, and cannot be run for sample sizes much larger than $n=10\,000$. Efficient approximations have been developed allowing to run approximate t-SNE for much larger sample sizes \cite{maaten2014accelerating,linderman2017efficient}, but until now have only been implemented for $\nu=1$. As a result, the effect of $\nu\ne 1$ on large real-life datasets has remained unknown.

Here we show that the recent FIt-SNE approximation \cite{linderman2017efficient} can be modified to use an arbitrary value of $\nu$ and demonstrate that $\nu<1$ can reveal ``hidden'' structure, invisible with standard t-SNE.

\section{Results}

\subsection{t-SNE with arbitrary degree of freedom}

SNE defines directional affinity of point $\mathbf x_j$ to point $\mathbf x_i$ as
$$p_{j|i} = \frac{\exp(-\|\mathbf x_i - \mathbf x_j \|^2 / 2\sigma_i^2)}{\sum_{k\ne i} \exp(-\|\mathbf x_i - \mathbf x_k \|^2 / 2\sigma_i^2)}.$$ For each $i$, this forms a probability distribution over all points $j\ne i$ (all $p_{i|i}$ are set to zero). The variance of the Gaussian kernel $\sigma_i^2$ is chosen such that the \textit{perplexity} of this probability distribution $$\exp\Big(- \ln(2) \cdot \sum_{j\ne i} p_{j|i} \log_2 p_{j|i}\Big)$$ has some pre-specified value. In symmetric SNE (SSNE)\footnote{In the following text we will not make a distinction between the symmetric SNE (SSNE) and the original, asymmetric, SNE.} and t-SNE the affinities are symmetrised and normalised
$$p_{ij} = \frac{p_{i|j} + p_{j|i}}{2n}$$
to form a probability distribution on the set of all pairs $(i,j)$.

The points are then arranged in a low-dimensional space to minimise the Kullback-Leibler (KL) divergence between $p_{ij}$ and the affinities in the low-dimensional space, $q_{ij}$:
\begin{gather*}
\mathcal L = \sum_{i,j} p_{ij}\log\frac{p_{ij}}{q_{ij}}, \\
q_{ij} = \frac{w_{ij}}{Z},\;\;\;w_{ij} = k(\|\mathbf y_i - \mathbf y_j\|),\;\;\; Z=\sum_{k\ne l} w_{kl}.
\end{gather*}
Here $k(d)$ is a kernel that transforms Euclidean distance $d$ between any two points into affinities, and $\mathbf y_i$ are low-dimensional coordinates (all $q_{ii}$ are set to 0).

SNE uses the Gaussian kernel $k(d) = \exp(-d^2)$. T-SNE uses the t-distribution with one degree of freedom (also known as Cauchy distribution): $k(d) = 1/(1+d^2)$. Here we consider a general t-distribution kernel
\begin{equation}
k(d) = \frac{1}{(1+d^2/\nu)^{(\nu+1)/2}}.
\label{t-kernel}\tag{$\star$}
\end{equation}
As in \cite{yang2009heavy}, we use a simplified version defined as \begin{equation}
k(d) = \frac{1}{(1+d^2/\alpha)^\alpha}.
\label{our-kernel}\tag{$\star\star$}
\end{equation}
This kernel corresponds to the \textit{scaled} t-distribution with $\nu=2\alpha-1$. This means that using (\ref{our-kernel}) instead of (\ref{t-kernel}) in t-SNE produces an identical output apart from the global scaling by $\sqrt{2\nu/(\nu+1)}$. At the same time, (\ref{our-kernel}) allows to use any $\alpha>0$, including $\alpha \in (0,1/2]$ corresponding to negative $\nu$, i.e. it allows kernels with tails heavier than any possible t-distribution.\footnote{Equivalently, we could use an even simpler kernel $k(d)=(1+d^2)^{-\alpha}$ that differs from (\ref{our-kernel}) only by scaling. We prefer (\ref{our-kernel}) because of the explicit Gaussian limit at $\alpha\to\infty$.} Yang et al. \cite{yang2009heavy} use the same kernel but with $\alpha$ replaced by $1/\alpha$, and call it ``heavy-tailed SNE'' (HSSNE).

The gradient of the loss function (see Appendix or \cite{yang2009heavy}) is
$$\frac{\partial \mathcal L}{\partial \mathbf y_i} = 4\sum_j (p_{ij}-q_{ij})w_{ij}^{1/\alpha}(\mathbf y_i- \mathbf y_j).$$
Any implementation of exact t-SNE can be easily modified to use this expression instead of the $\alpha=1$ gradient.

Modern t-SNE implementations make two approximations. First, they set most $p_{ij}$ to zero, apart from only a small number of close neighbours \cite{maaten2014accelerating,linderman2017efficient}, accelerating the attractive force computations (that can be very efficiently parallelised). This carries over to the $\alpha\ne 1$ case. The repulsive forces are approximated in FIt-SNE by interpolation on a grid, further accelerated with the Fourier transform \cite{linderman2017efficient}. This interpolation can be carried out for the $\alpha\ne 1$ case in full analogy to the $\alpha=1$ case (see Appendix).

Importantly, the runtime of FIt-SNE with $\alpha\ne 1$ is practically the same as with $\alpha=1$. For example, embedding MNIST ($n=70\,000$) with perplexity 50 as described below took 90~seconds with $\alpha=1$ and 97~seconds with $\alpha=0.5$ on a computer with 4 double-threaded cores, 3.4~GHz each.\footnote{The numbers correspond to 1000 gradient descent iterations. The slight speed decrease is due to a more efficient implementation of the interpolation code for the special case of $\alpha=1$.}

\subsection{Toy examples}

We first applied exact t-SNE with various values of $\alpha$ to a simple toy data set consisting of several well-separated clusters. Specifically, we generated a 10-dimensional data set with 100 data points in each of the 10 classes (1000 points overall). The points in class $i$ were sampled from a Gaussian distribution with covariance $\mathbf I_{10}$ and mean $\boldsymbol \mu_i = 4\mathbf e_i$ where $\mathbf e_i$ is the $i$-th basis vector. We used perplexity 50, and default optimisation parameters (1000 iterations, learning rate 200, early exaggeration 12, length of early exaggeration 250, initial momentum 0.5, switching to 0.8 after 250 iterations).

\begin{figure*}[t]
\includegraphics[width=\textwidth]{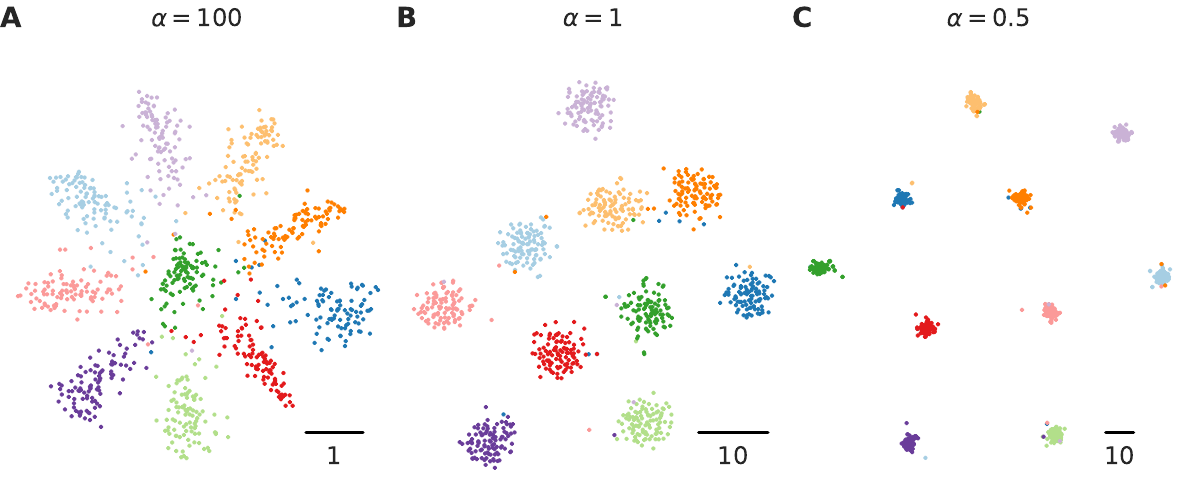}
\caption{Toy example with ten Gaussian clusters. \textbf{(A)} SNE visualisation of 10 spherical clusters that are all equally far away from each other ($\alpha=100$). \textbf{(B)} Standard t-SNE visualisation of the same data set ($\alpha=1$). \textbf{(C)} t-SNE visualisation with $\alpha=0.5$. The same random seed was used for initialisation in all panels. Scale bars are shown in the bottom-right of each panel.}
\label{fig-toy1}
\end{figure*}

Figure~\ref{fig-toy1} shows the t-SNE results for $\alpha=100$, $\alpha=1$, and $\alpha=0.1$. A t-distribution with $\nu=2\alpha-1=199$ degrees of freedom is very close to the Gaussian distribution, so here and below we will refer to the $\alpha=100$ result as SNE. We see that class separation monotonically increases with decreasing $\alpha$: t-SNE (Figure~\ref{fig-toy1}B) separates the classes much better than SNE (Figure~\ref{fig-toy1}A), but t-SNE with $\alpha=0.5$ separates them much better still (Figure~\ref{fig-toy1}C).

\begin{figure*}[t]
\includegraphics[width=\textwidth]{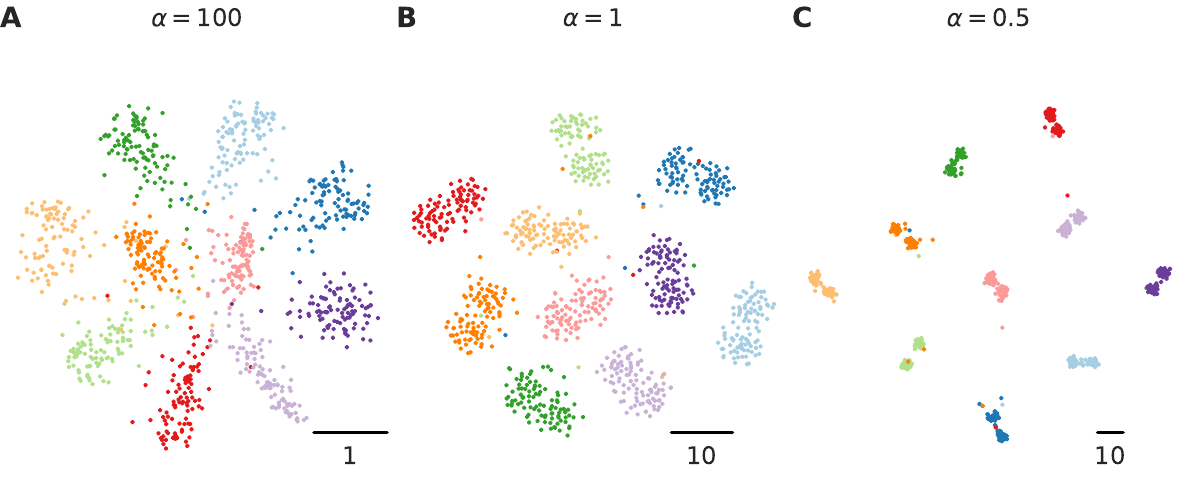}
\caption{Toy example with ten ``dumbbell''-shaped clusters. \textbf{(A)} SNE visualisation of 10 dumbbell-shaped clusters ($\alpha=100$). \textbf{(B)} Standard t-SNE visualisation ($\alpha=1$). \textbf{(C)} t-SNE visualisation with $\alpha=0.5$.}
\label{fig-toy2}
\end{figure*}

In the above toy example, the choice between different values of $\alpha$ is mostly aesthetic. This is not the case in the next toy example. Here we change the dimensionality to 20 and shift 50 points in each class by $2\mathbf e_{10+i}$ and the remaining 50 points by $-2\mathbf e_{10+i}$ (where $i$ is the class number). The intuition is that now each of the 10 classes has a ``dumbbell'' shape. This shape is invisible in SNE (Figure~\ref{fig-toy2}A) and hardly visible in standard t-SNE (Figure~\ref{fig-toy2}B), but becomes apparent with $\alpha=0.5$ (Figure~\ref{fig-toy2}C). In this case, decreasing $\alpha$ below 1 is necessary to bring out the fine structure of the data.

\subsection{Mathematical analysis}

We showed that decreasing $\alpha$ increases cluster separation (Figures~\ref{fig-toy1}, \ref{fig-toy2}). Why does this happen? An informal argument is that in order to match the between-cluster affinities $p_{ij}$, the distance between clusters in the t-SNE embedding needs to grow when the kernel becomes progressively more heavy-tailed \cite{maaten2008visualizing}.

\begin{figure}[t]
\begin{center}
\includegraphics[width=.6\textwidth]{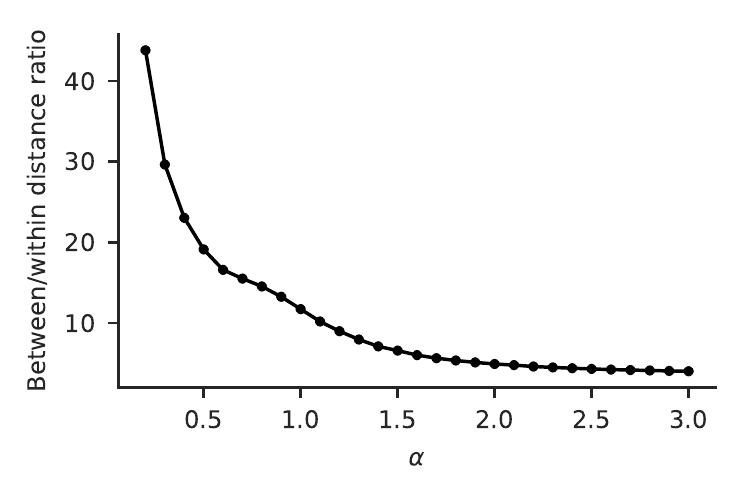}
\caption{Separation in the t-SNE visualisation between the two well-separated clusters as a function of $\alpha$. Separation was measured as the between-centroids distance divided by the root-mean-square within-cluster distance.}
\label{fig-separation}
\end{center}
\end{figure}

To quantify this effect, we consider an example of two standard Gaussian clusters in 10 dimensions ($n=100$ in each) with the between-centroid distance set to $5\sqrt{2}$; these clusters can be unambiguously separated. We use exact t-SNE (perplexity 50) with various values of $\alpha$ from 0.2 to 3.0 and measure the cluster separation in the embedding. As a scale-invariant measure of separation we used between-centroids distance divided by the root-mean-square within-cluster distance. Indeed, we observed a monotonic decrease of this measure with growing $\alpha$ (Figure~\ref{fig-separation}).

The informal argument mentioned above can be replaced by the following formal one. Consider two high-dimensional clusters ($n$ points in each) with all pairwise within-cluster distances equal to $D_w$ and all pairwise between-cluster distances equal to $D_b\gg D_w$ (this can be achieved in the space of $2n$ dimensions). In this case, the $p_{ij}$ matrix has only two unique non-zero values: all within-cluster affinities are given by $p_w$ and all between-cluster affinities by $p_b$, 
\begin{align*}
p_w &= \frac{K(D_w)}{n\big[(n-1)K(D_w) + nK(D_b)\big]}\\
p_b &= \frac{K(D_b)}{n\big[(n-1)K(D_w) + nK(D_b)\big]},
\end{align*}
where $K(D)$ is the Gaussian kernel corresponding to the chosen perplexity value. Consider an exact t-SNE mapping to the space of the same dimensionality. In this idealised case, t-SNE can achieve zero loss by setting within- and between-cluster distances $d_w$ and $d_b$ in the embedding such that $q_w = p_w$ and $q_b = p_b$. This will happen if
$$\frac{k(d_b)}{k(d_w)} = \frac{K(D_b)}{K(D_w)}.$$ Plugging in the expression for $k(d)$ and denoting the constant right-hand side by $c<1$, we obtain
$$\sqrt{\frac{\alpha + d_b^2}{\alpha + d_w^2}} = c^{-1/(2\alpha)}.$$
The left-hand side can be seen as a measure of class separation close to the one used in Figure~\ref{fig-separation}, and the right-hand side monotonically decreases with increasing $\alpha$.

In the simulation shown in Figure~\ref{fig-separation}, the $p_{ij}$ matrix does not have only two unique elements, the target dimensionality is two, and the t-SNE cannot possibly achieve zero loss. Still, qualitatively we observe the same behaviour: approximately power-law decrease of separation with increasing $\alpha$.

\subsection{Real-life data sets}

We now demonstrate that these theoretical insights are relevant to practical use cases on large-scale data sets. Here we use approximate t-SNE (FIt-SNE).

\subsubsection{MNIST}

\begin{figure*}[t]
\includegraphics[width=\textwidth]{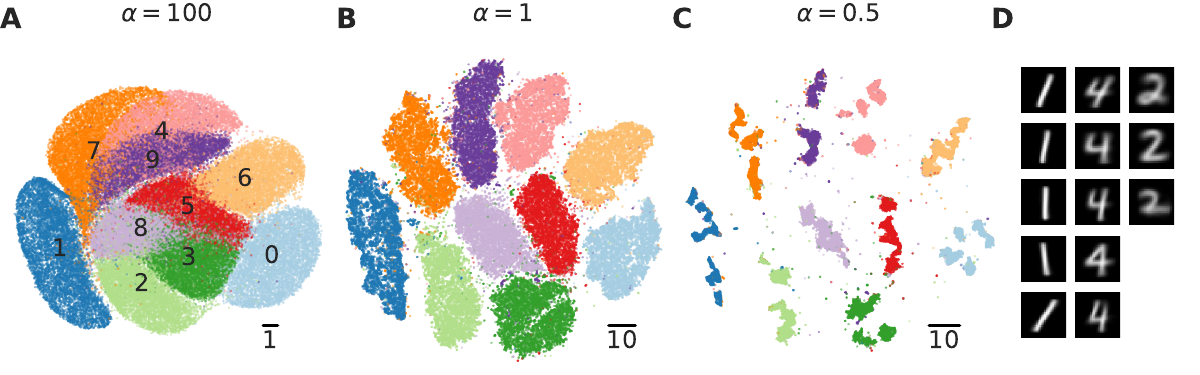}
\caption{MNIST data set ($n=70\,000)$. \textbf{(A)} SNE visualisation ($\alpha=100$). \textbf{(B)} Standard t-SNE visualisation ($\alpha=1$). \textbf{(C)} t-SNE visualisation with $\alpha=0.5$. The colours are consistent across panels (A--C), labels are shown in (A). PCA initialisation was used in all three cases. Transparency 0.5 for all dots in all three panels. \textbf{(D)} Average images for some individual sub-clusters from (C). The sub-clusters were isolated via DBSCAN with default settings as it is implemented in \texttt{scikit-learn}. Up to five subclusters with at least 100 points are shown, ordered from top to bottom by abundance.}
\label{fig-mnist}
\end{figure*}

We applied t-SNE with various values of $\alpha$ to the MNIST data set (Figure~\ref{fig-mnist}), comprising $n=70\,000$ grayscale $28\times 28$ images of handwritten digits. As a pre-processing step, we used principal component analysis (PCA) to reduce the dimensionality from 784 to 50. We used perplexity 50 and default optimisation parameters apart from learning rate that we increased to $\eta=1000$.\footnote{To get a good t-SNE visualisation of MNIST, it is helpful to increase either the learning rate or the length of the early exaggeration phase. Default optimisation parameters often lead to some of the digits being split into two clusters. In the cytometric context, this phenomenon was described in detail by \cite{belkina2018automated}.} For easier reproducibility, we initialised the t-SNE embedding with the first two PCs (scaled such that PC1 had standard deviation 0.0001). 

To the best of our knowledge, Figure~\ref{fig-mnist}A is the first existing SNE ($\alpha=100$) visualisation of the whole MNIST: we are not aware of any SNE implementation that can handle a dataset of this size. It produces a surprisingly good visualisation but is nevertheless clearly outperformed by standard t-SNE ($\alpha=1$, Figure~\ref{fig-mnist}B): many digits coalesce together in SNE but get separated into clearly distinct clusters in t-SNE. Remarkably, reducing $\alpha$ to 0.5 makes each digit further split into multiple separate sub-clusters (Figure~\ref{fig-mnist}C), revealing a fine structure within each of the digits.

To demonstrate that these sub-clusters are meaningful, we computed the average MNIST image for some of the sub-clusters (Figure~\ref{fig-mnist}D). In each case, the shapes appear to be meaningfully distinct: e.g. for the digit ``4'', the hand-writing is more italic in one sub-cluster, more wide in another, and features a non-trivial homotopy group (i.e. has a loop) in yet another one. Similarly, digit ``2'' is separated into three sub-clusters, with the most abundant one showing a loop in the bottom-left and the next abundant one having a sharp angle instead. Digit ``1'' is split according to the stroke angle. Re-running t-SNE using random initialisation with different seeds yielded consistent results. Points that appear as outliers in Figure~\ref{fig-mnist}C mostly correspond to confusingly written digits.

MNIST has been a standard example for t-SNE starting from the original t-SNE paper \cite{maaten2008visualizing}, and it has been often observed that t-SNE preserves meaningful within-digit structure. Indeed, the sub-clusters that we identified in Figure~\ref{fig-mnist}C are usually close together in Figure~\ref{fig-mnist}B.\footnote{This can be clearly seen in an animation that slowly decreases $\alpha$ from 100 to 0.5, see \url{http://github.com/berenslab/finer-tsne}.} However, standard t-SNE does not separate them into visually isolated sub-clusters, and so does not make this internal structure obvious.

\subsubsection{Single-cell RNA-sequencing data}

\begin{figure*}[ht]
\includegraphics[width=\textwidth]{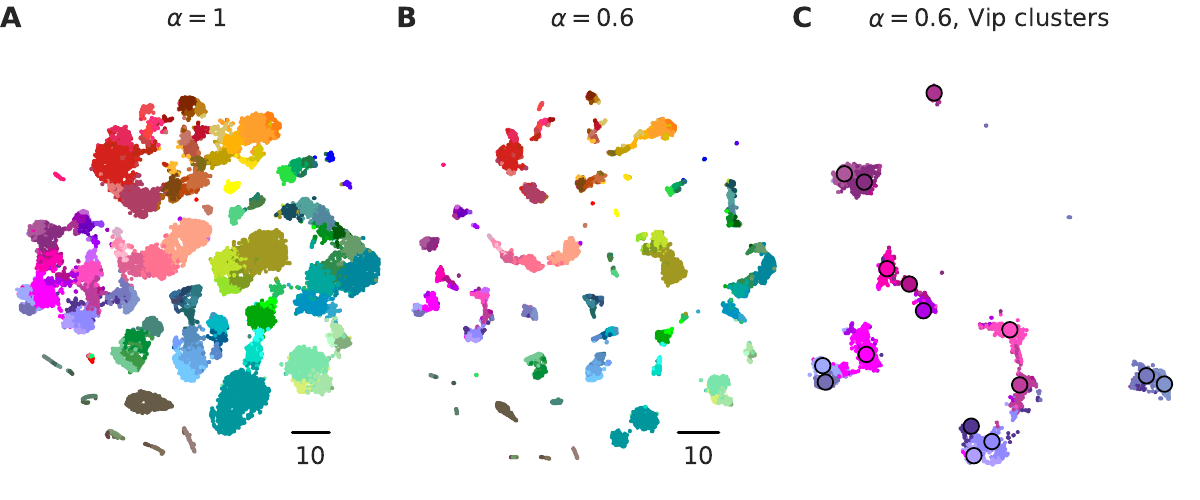}
\caption{Tasic et al. data set ($n=23\,822$). \textbf{(A)} Standard t-SNE visualisation ($\alpha=1$). Cluster ids and cluster colours are taken from the original publication \cite{tasic2018shared}: cold colours for excitatory neurons, warm colours for inhibitory neurons, and grey/brown colours for non-neural cells such as
astrocytes or microglia. \textbf{(B)} t-SNE visualisation with $\alpha=0.6$. \textbf{(C)} A zoom-in into the left side of panel (B) showing all \textit{Vip} clusters from Tasic et al. Black circles mark cluster centroids (medians).}
\label{fig-tasic}
\end{figure*}

For the second example, we took the transcriptomic dataset from \cite{tasic2018shared}, comprising $n=23\,822$ cells from adult mouse cortex (sequenced with Smart-seq2 protocol). Dimensions are genes, and the data are the integer counts of RNA transcripts of each gene in each cell. Using a custom expert-validated clustering procedure, the authors divided these cells into 133 clusters. In Figure~\ref{fig-tasic}, we used the cluster ids and cluster colours from the original publication. 

Figure~\ref{fig-tasic}A shows the standard t-SNE ($\alpha=1$) of this data set, following common transcriptomic pre-processing steps as described in \cite{kobak2018art}. Briefly, we  row-normalised and log-transformed the data, selected 3000 most variable genes and used PCA to further reduce dimensionality to 50. We used perplexity 50 and PCA initialisation. The resulting t-SNE visualisation is in a reasonable agreement with the clustering results, however it lumps many clusters together into contiguous ``islands'' or ``continents'' and overall suggests many fewer than 133 distinct clusters.

Reducing the number of degrees of freedom to $\alpha=0.6$ splits many of the contiguous islands into ``archipelagos'' of smaller disjoint areas (Figure~\ref{fig-tasic}B). In many cases, this roughly agrees with the clustering results of \cite{tasic2018shared}. Figure~\ref{fig-tasic}C shows a zoom-in into the \textit{Vip} clusters (west-southwest part of panel B) that provide one such example: isolated islands correspond well to the individual clusters (or sometimes pairs of clusters). Importantly, the cluster labels in this data set are not ground truth; nevertheless the agreement between cluster labels and t-SNE with $\alpha=0.6$ provides additional evidence that this data categorisation is meaningful.

\subsubsection{HathiTrust library}

For the final example, we used the HathiTrust library data set \cite{schmidt2018stable}. The full data set comprises 13.6 million books and can be described with several million features that represent word counts of each word in each book. We used the pre-processed data from \cite{schmidt2018stable}: briefly, the  word counts were row-normalised, log-transformed, projected to 1280 dimensions using random linear projection with coefficients $\pm1$, and then reduced to 100 PCs.\footnote{The $13.6\cdot 10^6 \times 100$ data set was downloaded from \url{https://zenodo.org/record/1477018}.} The available meta-data include author name, book title, publication year, language, and Library of Congress classification (LCC) code. For simplicity, we took a $n=408\,291$ subset consisting of all books in Russian language. We used perplexity 50 and learning rate $\eta=10\,000$.

\begin{figure*}[ht]
\includegraphics[width=\textwidth]{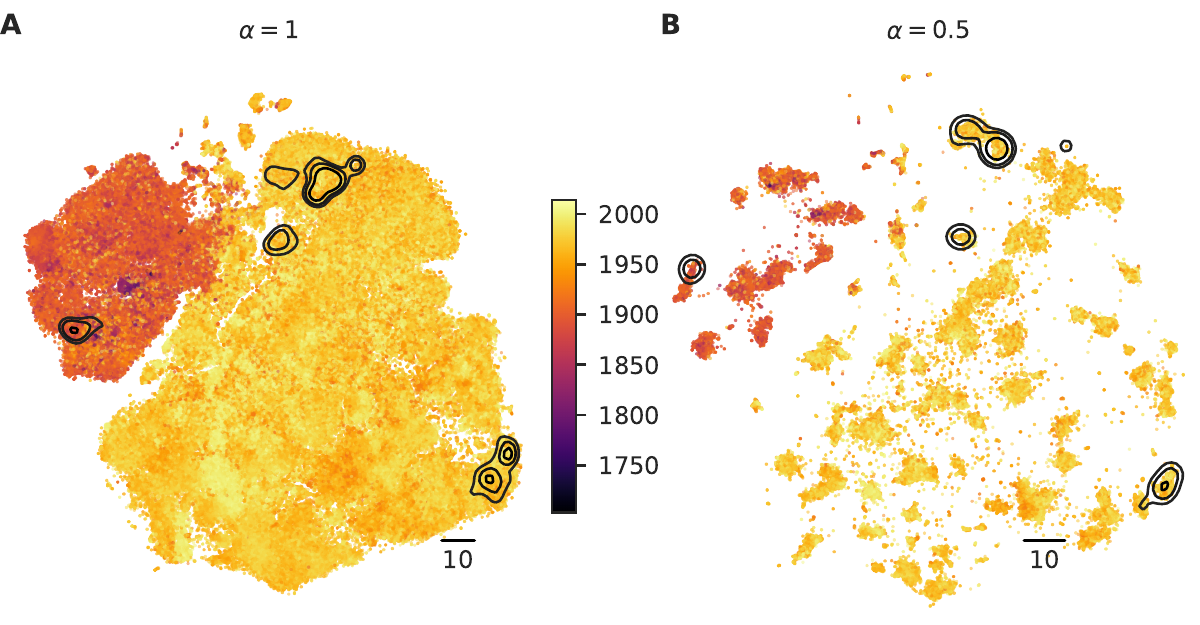}
\caption{Russian language part of the HathiTrust library ($n=408\,291$). \textbf{(A)} Standard t-SNE visualisation ($\alpha=1$). Colour denotes publication year. \textbf{(B)} t-SNE visualisation with $\alpha=0.5$. Black contours in both panels are kernel density estimate contour lines for mathematical literature (lower right) and poetry (upper left),  plotted with \texttt{seaborn.kdeplot()} with Gaussian bandwidth set to 2.0. Contour levels were manually tuned to enclose the majority of the books).}
\label{fig-hathi}
\end{figure*}

Figure~\ref{fig-hathi}A shows the standard t-SNE visualisation ($\alpha=1$) coloured by the publication year. The most salient feature is that pre-1917 books cluster together (orange/red colours): this is due to the major reform of Russian orthography implemented in 1917, leading to most words changing their spelling. However, not much of a substructure can be seen among the books published after (or before) 1917. In contrast, t-SNE visualisation with $\alpha=0.5$ fragments the corpus into a large number of islands (Figure~\ref{fig-hathi}B).

We can identify some of the islands by inspecting the available meta-data. For example, mathematical literature (LCC code \texttt{QA}, $n=6490$ books) is not separated from the rest in standard t-SNE, but occupies the leftmost island in t-SNE with $\alpha=0.5$ (blue contour lines in both panels). Several neighbouring islands correspond to the physics literature (LCC code \texttt{QC}, $n=5104$ books; not shown). In an attempt to capture something radically different from mathematics, we selected all books authored by several famous Russian poets\footnote{Anna Akhmatova, Alexander Blok, Joseph Brodsky, Afanasy Fet, Osip Mandelstam, Vladimir Mayakovsky, Alexander Pushkin, and Fyodor Tyutchev.} ($n=1369$ in total). This is not a curated list: there are non-poetry books authored by these authors, while many other poets were not included (the list of poets was not cherry-picked; we made the list before looking at the data). Nevertheless, when using $\alpha=0.5$, the poetry books printed after 1917 seemed to occupy two neighbouring islands, and the ones printed before 1917 were reasonably isolated as well (Figure~\ref{fig-hathi}B, black contour lines). In the standard t-SNE visualisation poetry was not at all separated from the surrounding population of books.

\section{Related work}

Yang et al. \cite{yang2009heavy}  introduced symmetric SNE with the kernel family $$k(d) = \frac{1}{(1+\alpha d^2)^{1/\alpha}},$$ calling it ``heavy-tailed symmetric SNE'' (HSSNE). This is exactly the same kernel family as (\ref{our-kernel}), but with $\alpha$ replaced by $1/\alpha$. However, Yang et al. did not show any examples of heavier-tailed kernels revealing additional structure compared to $\alpha=1$ and did not provide an implementation suitable for large sample sizes. Interestingly, Yang et al. argued that gradient descent is not suitable for HSSNE and suggested an alternative optimisation algorithm; here we demonstrated that the standard t-SNE optimisation works reasonably well in a wide range of $\alpha$ values (but see Discussion).

Van der Maaten \cite{maaten2009learning} discussed the choice of the degree of freedom in the t-distribution kernel in the context of parametric t-SNE. He argued that $\nu>1$ might be warranted when embedding the data in more than two dimensions. He also implemented a version of parametric t-SNE that optimises over $\nu$. However, similar to \cite{yang2009heavy}, \cite{maaten2009learning} did not contain any examples of $\nu<1$ being actually useful in practice.

UMAP \cite{mcinnes2018umap} is a promising recent algorithm closely related to an earlier largeVis \cite{tang2016visualizing}; both are similar to t-SNE but modify the repulsive forces to make them amenable for a sampling-based stochastic optimisation. UMAP uses the following family of similarity kernels: $$k(d) = \frac{1}{1+ad^{2b}},$$ which reduces to Cauchy when $a=b=1$ and is more heavy-tailed when $0<b<1$. UMAP default is $a\approx 1.6$ and $b\approx 0.9$ with both parameters adjusted via the \texttt{min\_dist} input parameter (default value 0.1). Decreasing \texttt{min\_dist} all the way to zero corresponds to decreasing $b$ to 0.79. In our experiments, we observed that modifying \texttt{min\_dist} (or $b$ directly) led to an effect qualitatively similar to modifying $\alpha$ in t-SNE. For some data sets this required manually decreasing $b$ below 0.79. In case of MNIST, $b=0.3$, but not $b=0.79$, revealed  sub-digit structure (Figure~\ref{fig-umap}) --- an effect that has not been described before (cf. \cite{mcinnes2018umap} where McInnes et al. state that \texttt{min\_dist} is ``an essentially aesthetic parameter''). In other words, the same conclusion seems to apply to UMAP: heavy-tailed kernels reveal a finer cluster structure. A more in-depth study of the relationships between the two algorithms is beyond the scope of this paper. 

\section{Discussion}

We showed that using $\alpha<1$ in t-SNE can yield insightful visualisations that are qualitatively different compared to the standard choice of $\alpha=1$. Crucially, the choice of $\alpha=1$ was made in \cite{maaten2008visualizing} for the reasons of mathematical convenience, and we are not aware of any \textit{a priori} argument in favour of $\alpha=1$. As $\alpha\ne 1$ still yields a t-distribution kernel (scaled t-distribution to be precise), we prefer not to use a separate acronym (HSSNE \cite{yang2009heavy}). If needed, one can refer to t-SNE with $\alpha<1$ as ``heavy-tailed'' t-SNE.

We found that lowering $\alpha$ below 1 makes progressively finer structure apparent in the visualisation and brings out smaller clusters, which --- at least in the data sets studied here --- are often meaningful. In a way, $\alpha<1$ can be thought of as a ``magnifying glass'' for the standard t-SNE representation. We do not think that there is one ideal value of $\alpha$ suitable for all data sets and all situations; instead we consider it a useful adjustable parameter of t-SNE, complementary to the perplexity. We observed a non-trivial interaction between $\alpha$ and perplexity: Small vs. large perplexity makes the affinity matrix $p_{ij}$ represent the local vs. global structure of the data \cite{kobak2018art}. Small vs. large $\alpha$ makes the embedding represent the finer vs. coarser structure of the affinity matrix. In practice, it can make sense to treat it as a two-dimensional parameter space to explore. However, for large data sets ($n\gtrsim 10^6$), it is computationally unfeasible to substantially increase the perplexity from its standard range of 30--100 (as it would prohibitively increase the runtime), and so $\alpha$ becomes the only available parameter to adjust. 

\begin{figure*}[t]
\begin{center}
\includegraphics[width=.7\textwidth]{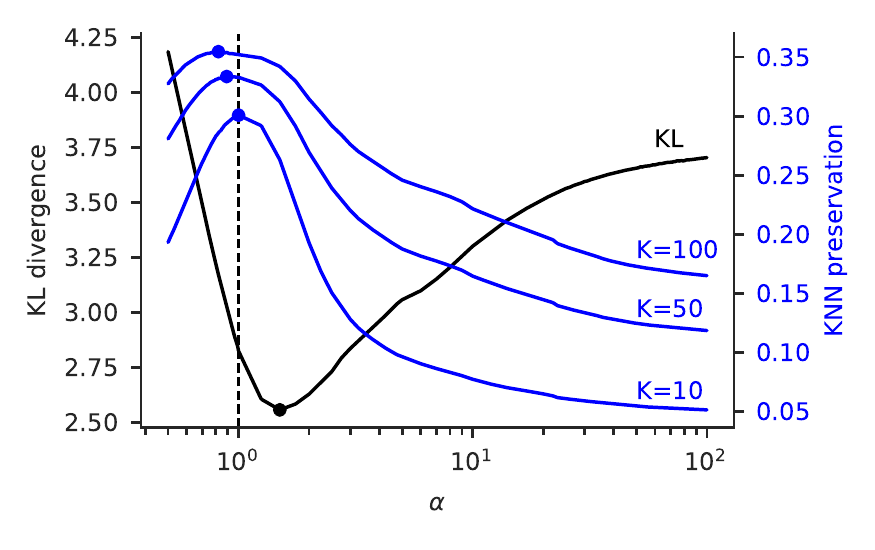}
\caption{Quality assessment of the MNIST embedding with $\alpha\in[0.5, 100]$ after 1000 gradient descent iterations with learning rate $\eta=1000$ (scaled PCA initialisation). The horizontal axis is on the log scale. The $\alpha$ values were sampled on a grid with step $0.01$ for $\alpha<1$, $0.25$ for $1\le\alpha\le5$ and $1$ for $\alpha>5$. The black line shows KL divergence (left axis) with minimum at $\alpha=1.5$. Running gradient descent with $\alpha=0.5$ for $10\,000$ iterations (Figure~\ref{fig-mnist10k}) lowered KL divergence down to 3.6, which was still above the minimum value. Blue lines show neighbourhood preservation (the fraction of $k$ nearest neighbours of each point that remain within $k$ nearest neighbours in the embedding, averaged over all $n=70\,000$ points) for $k=10$, $k=50$, and $k=100$.}
\label{fig-kl}
\end{center}
\end{figure*}

One important caveat is to be kept in mind. It is well-known that t-SNE, especially with low perplexity, can find ``clusters'' in pure noise, picking up random fluctuations in the density \cite{wattenberg2016use}. This can happen with $\alpha=1$ but gets exacerbated with lower values of $\alpha$. A related point concerns clustered real-life data where separate clusters (local density peaks) can sometimes be connected by an area of lower but non-zero density: for example, \cite{tasic2018shared} argued that many pairs of their 133 clusters have intermediate cells. Our experiments demonstrate that lowering $\alpha$ can make such clusters more and more isolated in the embedding, creating a potentially misleading appearance of perfect separation (see e.g. Figure~\ref{fig-toy1}). In other words, there is a trade-off between bringing out finer cluster structure and preserving continuities between clusters.

Choosing a value of $\alpha$ that yields the most faithful representation of a given data set is challenging because it is difficult to quantify ``faithfulness'' of any given embedding \cite{lee2009quality}. For example, for MNIST, KL divergence is minimised at $\alpha\approx 1.5$ (Figure~\ref{fig-kl}), but it may not be the ideal metric to quantify the embedding quality \cite{im2018stochastic}. Indeed, we found that $k$-nearest neighbour (KNN) preservation \cite{lee2009quality} peaked elsewhere: the peak for $k=10$ was at $\alpha\approx 1.0$,  for $k=50$ at $\alpha\approx 0.9$, and for $k=100$ at $\alpha\approx 0.8$ (Figure~\ref{fig-kl}). We stress that we do not think that KNN preservation is the most appropriate metric here; our point is that different metrics can easily disagree with each other. In general, there may not be a single ``best'' embedding of high-dimensional data in a two-dimensional space. Rather, by varying $\alpha$, one can obtain different complementary ``views'' of the data.

Very low values of $\alpha$ correspond to kernels with very wide and very flat tails, leading to vanishing gradients and difficult convergence. We found that $\alpha=0.5$ was about the smallest value that could be safely used (Figure~\ref{fig-lowalpha}). In fact, it may take more iterations to reach convergence for $0.5<\alpha<1$ compared to $\alpha=1$. As an example, running t-SNE on MNIST with $\alpha=0.5$ for ten times longer than we did for Figure~\ref{fig-mnist}C, led to the embedding expanding much further (which leads to a slow-down of FIt-SNE interpolation) and, as a result, resolving additional sub-clusters (Figure~\ref{fig-mnist10k}). On a related note, when using only one single MNIST digit as an input for t-SNE with $\alpha=0.5$, the embedding also fragments into many more clusters (Figure~\ref{fig-isolateddigits}), which we hypothesise is due to the points rapidly expanding to occupy a much larger area compared to what happens in the full MNIST embedding (Figure~\ref{fig-isolateddigits}). This can be counterbalanced by increasing the strength of the attractive forces (Figure~\ref{fig-isolateddigits}). Overall, the effect of the embedding scale on the cluster resolution remains an open research question.

In conclusion, we have shown that adjusting the heaviness of the kernel tails in t-SNE can be a valuable tool for data exploration and visualisation. As a practical recommendation, we suggest to embed any given data set using various values of $\alpha$, each inducing a different level of clustering, and hence providing insight that cannot be obtained from the standard $\alpha=1$ choice alone.\footnote{Our code is available at \url{http://github.com/berenslab/finer-tsne}. The main FIt-SNE repository at \url{http://github.com/klugerlab/FIt-SNE} was updated to support any $\alpha$ (version 1.1.0).}

\newpage
\section{Appendix}

The loss function, up to a constant term $\sum p_{ij}\log p_{ij}$, can be rewritten as follows:
\begin{align}
\mathcal L &= - \sum_{i,j} p_{ij}\log q_{ij} = -\sum_{i,j} p_{ij}\log \frac{w_{ij}}{Z} \nonumber \\
&=-\sum_{i,j} p_{ij}\log w_{ij} + \log \sum_{i,j}w_{ij}, \label{loss}
\end{align}
where we took into account that $\sum p_{ij}=1$. The first term in Eq. (\ref{loss}) contributes attractive forces to the gradient while the second term yields repulsive forces. The gradient is 
\begin{align}
\frac{\partial \mathcal L}{\partial \mathbf y_i} &=
-2\sum_j p_{ij} \frac{1}{w_{ij}} \frac{\partial w_{ij}}{\partial \mathbf y_i} + 2\sum_j \frac{1}{Z} \frac{\partial w_{ij}}{\partial \mathbf y_i} \label{gradient1}\\
&= -2\sum_j(p_{ij}-q_{ij})\frac{1}{w_{ij}}\frac{\partial w_{ij}}{\partial \mathbf y_i}. \label{gradient2}
\end{align}
The first expression is more convenient for numeric optimisation while the second one can be more convenient for mathematical analysis.

For the kernel $$k(d) = \frac{1}{(1+d^2/\alpha)^\alpha}$$
the gradient of $w_{ij} = k(\|\mathbf y_i - \mathbf y_j\|)$ is 
\begin{equation}
\frac{\partial w_{ij}}{\partial \mathbf y_i} = -2w^\frac{\alpha+1}\alpha (\mathbf y_i-\mathbf y_j). \label{w-gradient}
\end{equation}

Plugging Eq. \ref{w-gradient} into Eq. \ref{gradient2}, we obtain the expression for the gradient \cite{yang2009heavy}\footnote{Note that the C++ Barnes-Hut t-SNE implementation \cite{maaten2014accelerating} absorbed the factor 4 into the learning rate, and the FIt-SNE implementation \cite{linderman2017efficient} followed this convention.}
$$\frac{\partial \mathcal L}{\partial \mathbf y_i} =
4\sum_j (p_{ij}-q_{ij})w_{ij}^{1/\alpha}(\mathbf y_i-\mathbf y_j).$$

For numeric optimisation it is convenient to split this expression into the attractive and the repulsive terms. Plugging Eq. \ref{w-gradient} into Eq. \ref{gradient1}, we obtain
$$\frac{\partial \mathcal L}{\partial \mathbf y_i} = \mathbf F_\mathrm{att} + \mathbf F_\mathrm{rep}$$ where
\begin{align*}
\mathbf F_\mathrm{att} &= 4\sum_j p_{ij}w_{ij}^{1/\alpha} (\mathbf y_i-\mathbf y_j)\\ \mathbf F_\mathrm{rep} &= -4\sum_j w_{ij}^\frac{\alpha+1}{\alpha}/Z(\mathbf y_i-\mathbf y_j)
\end{align*}
It is noteworthy that the expression for $\mathbf F_\mathrm{attr}$ has $w_{ij}$ raised to the $1/\alpha$ power, which cancels out the fractional power in $k(d)$. This makes the runtime of $\mathbf F_\mathrm{attr}$ computation unaffected by the value of $\alpha$. In FIt-SNE, the sum over $j$ in $\mathbf F_\mathrm{attr}$ is approximated by the sum over $3\Pi$ approximate nearest neighbours of point $i$ obtained using Annoy \cite{annoy}, where $\Pi$ is the provided perplexity value. The $3\Pi$ heuristic comes from \cite{maaten2014accelerating}. The remaining $p_{ij}$ values are set to zero.

The $\mathbf F_\mathrm{rep}$ can be approximated using the interpolation scheme from \cite{linderman2017efficient}. It allows fast approximate computation of the sums of the form $$\textstyle{\sum}_j K(\|\mathbf y_i - \mathbf y_j\|)$$ and $$\textstyle{\sum}_j K(\|\mathbf y_i - \mathbf y_j\|)\mathbf y_j,$$ where $K(\cdot)$ is any smooth kernel, by using polynomial interpolation of $K$ on a fine grid.\footnote{The accuracy of in the interpolation can somewhat decrease for small values of $\alpha$. One can increase the accuracy by decreasing the spacing of the interpolation grid (see FIt-SNE documentation). We found that it did not noticeably affect the visualisations.} All kernels appearing in $\mathbf F_\mathrm{rep}$ are smooth.

\vspace{150px}
\section*{Acknowledgements}

This work was supported by the Deutsche Forschungsgemeinschaft (BE5601/4-1, EXC 2064, Project ID 390727645) (PB), the Federal Ministry of Education and Research (FKZ 01GQ1601, 01IS18052C), and the National Institute of Mental Health under award number U19MH114830 (DK and PB),  NIH grants F30HG010102 and U.S. NIH MSTP Training Grant T32GM007205 (GCL), NSF grant DMS-1763179 and the Alfred P. Sloan Foundation (SS), and the NIH grant R01HG008383 (YK). The content is solely the responsibility of the authors and does not necessarily represent the official views of the National Institutes of Health.

\newpage
\bibliography{main}
\bibliographystyle{splncs04}

\newpage
\onecolumn
\section*{Supplementary Figures}
\renewcommand{\thefigure}{S\arabic{figure}}
\setcounter{figure}{0}

\begin{figure*}[h!]
\includegraphics[width=\textwidth]{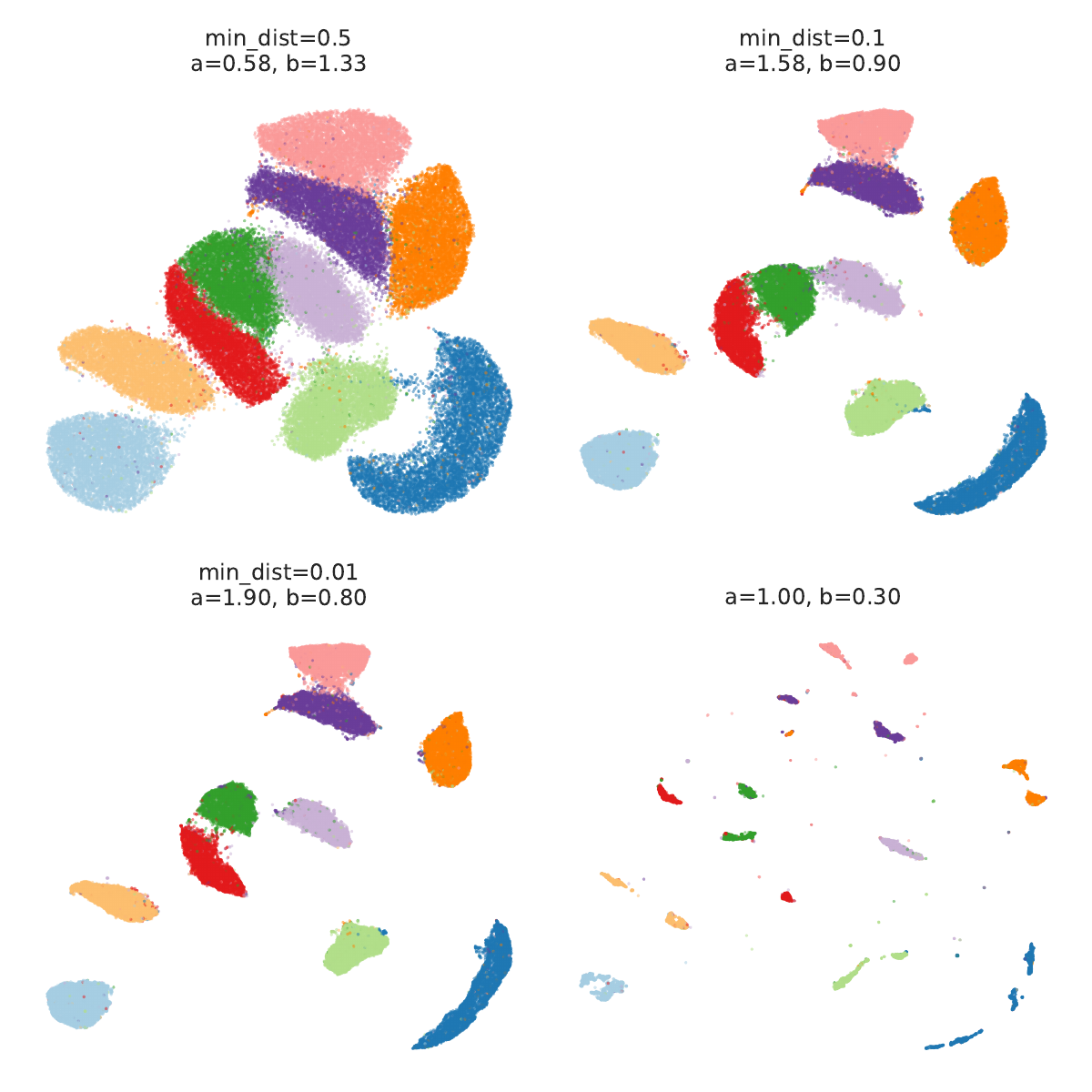}
\caption[several paragraphs]{UMAP visualisations of the MNIST data set with default parameters and \texttt{min\_dist} set to 0.5, 0.1 (default), 0.01, and $(a,b)$ manually set to $(1,0.3)$.  Each subplot shows the $a$ and $b$ values of the UMAP kernel $k(d)=(1+ad^{2b})^{-1}$. With \texttt{min\_dist} below 0.01 the $b$ as a function of \texttt{min\_dist} hardly changes. 

\setlength{\parindent}{0.5cm} As a side note, one can obtain an MNIST embedding very similar to the one that UMAP gives with default settings using t-SNE with late exaggeration of about $4$, i.e. multiplying all attractive forces by $4$ after the early exaggeration period (first 250 iterations) is over. This makes sense, given that the main difference between t-SNE and UMAP is in the repulsive term.}
\label{fig-umap}
\end{figure*}

\begin{figure*}[h!]
\includegraphics[width=\textwidth]{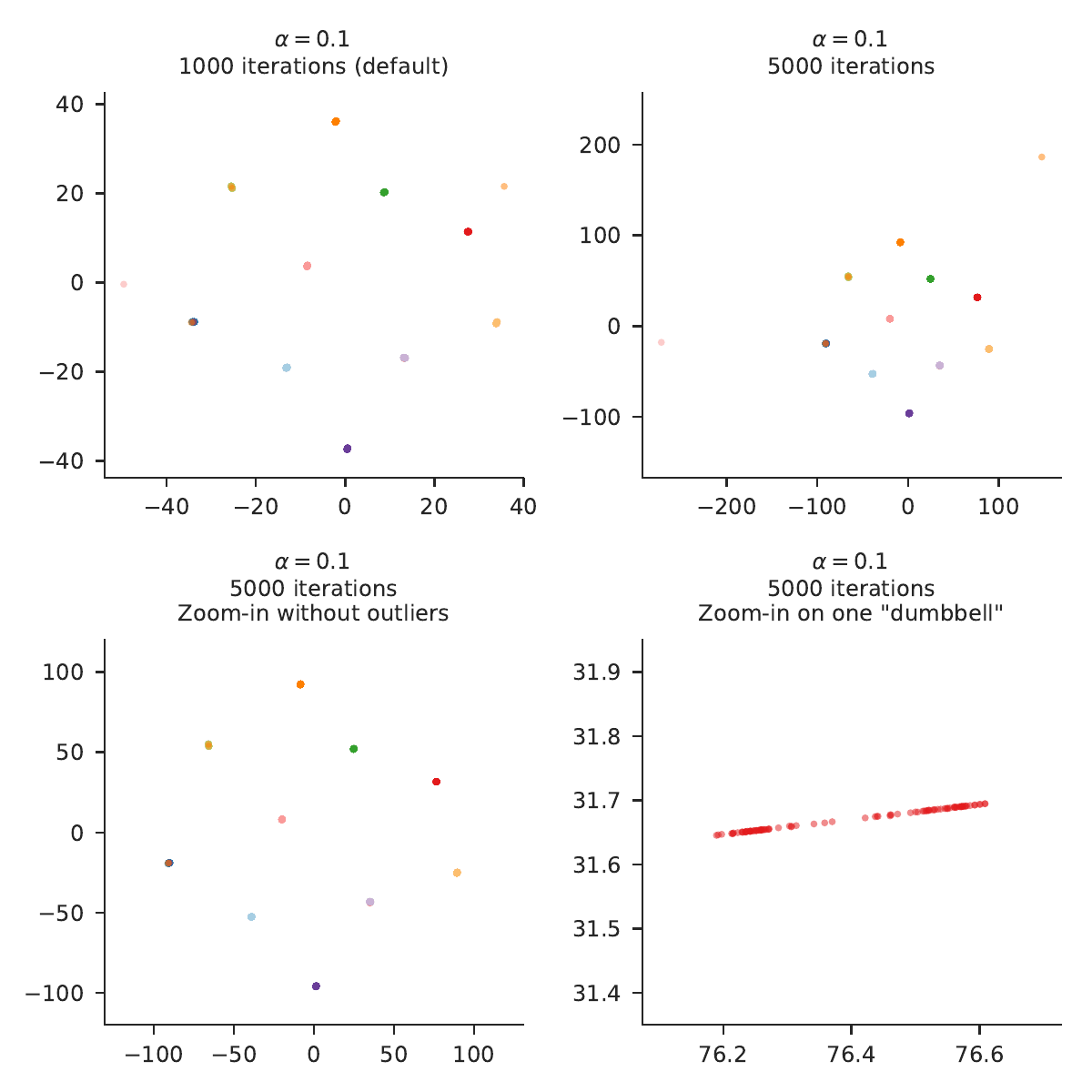}
\caption[several paragraphs]{Toy example with ten ``dumbbell''-shaped clusters from Figure~\ref{fig-toy2}, here embedded with $\alpha=0.1$. Top-left plot shows the result after 1000 gradient descent iterations (default). Note that the dumbbell shape is lost: whereas the number of visible clusters increased as $\alpha$ was lowered from 100 to $0.5$ (Figure~\ref{fig-toy2}), it decreased when it was further lowered to $0.1$. We believe the reason for this is that the strong repulsion between dumbbells ``squashes'' them in the beginning of optimisation into very compact blobs. It is likely that longer optimisation would resolve the dumbbell shapes. This is difficult to test because the kernel with $\alpha=0.1$ is extremely wide and flat, leading to slow convergence. Top-right plot shows the result after 5000 iterations. Here a few outlying points get pushed to the periphery. Zooming-in to the main 10 clusters (bottom-left) still does not resolve the dumbbell shapes. Further zooming in on one of the dumbbells (bottom-right) shows that the points are squashed into 1D which may be a sign of poor convergence.

\setlength{\parindent}{0.5cm} In a separate sets of experiments, we observed the similar phenomenon with MNIST: $\alpha=0.2$ after 1000 iterations yielded fewer clusters than $\alpha=0.5$. Our conclusion is that smaller values of $\alpha$ should be used with caution. }
\label{fig-lowalpha}
\end{figure*}

\begin{figure*}[h!]
\includegraphics[width=\textwidth]{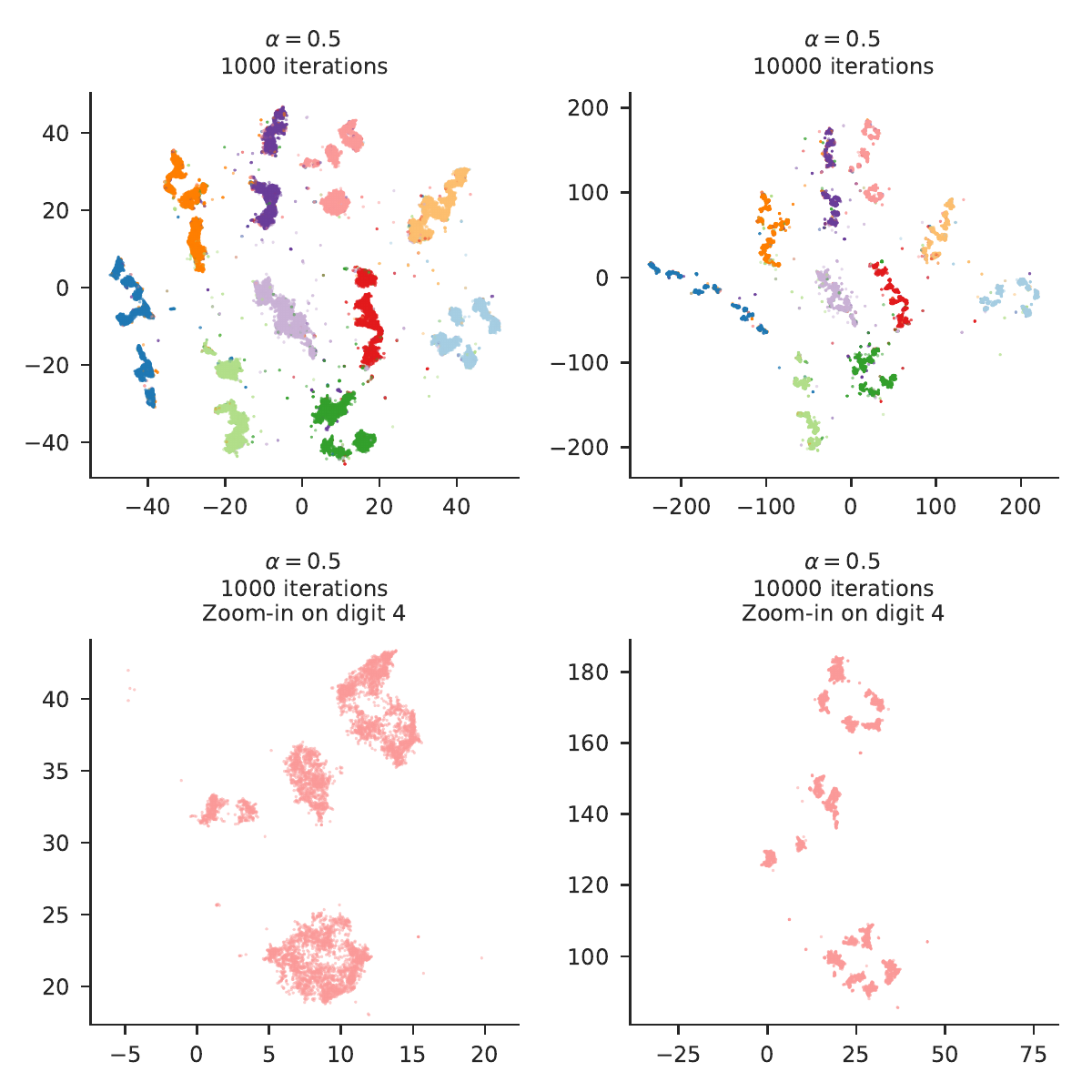}
\caption[several paragraphs]{t-SNE visualisation of the MNIST dataset with $\alpha=0.5$. The top-left panel is identical to Figure~\ref{fig-mnist}C; it was obtained with 1000 gradient descent iterations (the default value). The top-right panel corresponds to $10\,000$ iterations and has many more isolated sub-clusters. This can also be seen in the bottom row showing the respective zoom-ins into the digit ``4''. At the same time, the embedding after 1000 iterations is not misleading and is simply a coarser-grained version of the embedding after $10\,000$ iterations. 

\setlength{\parindent}{0.5cm} Using $10\,000$ iterations is impractical: whereas 1000 iterations were finished in 1.5 minutes, $10\,000$ iterations took 4 hours 30 minutes. This is because FIt-SNE interpolation scheme uses regular interpolation grid with the number of nodes growing quadratically with the embedding size. While the left embedding is contained within $[-50,50]^2$, the right one expands to $[-200,200]^2$. In principle, an implementation based on the fast multipole method (FMM) could be developed to dramatically accelerate the gradient computation in this setting where most of the embedding space is ``empty space'', but current FIt-SNE implementation does not support it.

Note that the standard t-SNE embedding with $\alpha=1$ also expands much further after $10\,000$ iterations, compared to the 1000 iterations. However, with $\alpha=1$ it does not resolve additional sub-clusters, at least in MNIST.}
\label{fig-mnist10k}
\end{figure*}

\begin{figure*}[h!]
\includegraphics[width=\textwidth]{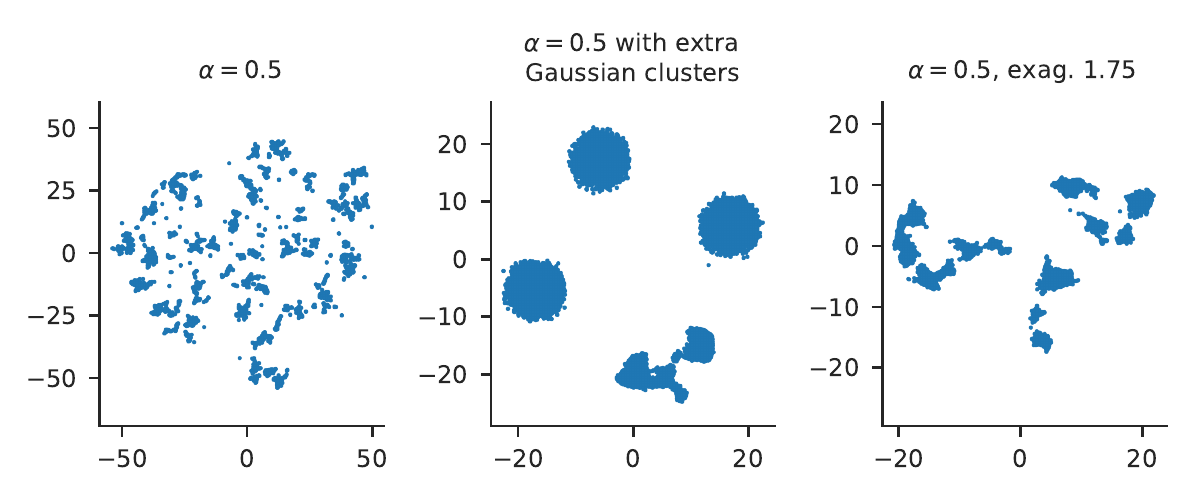}
\caption{t-SNE visualisation of a MNIST subset consisting of all images of digit ``4'' ($n=6\,824$) (perplexity 50).
Left: $\alpha=0.5$, the same as in Figure~\ref{fig-mnist}C. Note the large number of isolated clusters. We believe this happens because the embedding rapidly expands to a larger area, compared to Figure~\ref{fig-mnist10k} (bottom-left). One evidence for that is that re-running t-SNE after adding several random Gaussian clusters with $n=7\,000$ each, roughly recovers the shape of the digit ``4'' archipelago from the full MNIST embedding (middle). Right: $\alpha=0.5$ and exaggeration factor 1.5 \cite{kobak2018art}, i.e. all attractive forces are multiplied by 1.5 after the end of the early exaggeration phase (during the early exaggeration they are multiplied by 12). This roughly recovers the sub-clusters from the full MNIST embedding (Figure~\ref{fig-mnist10k}). The relationship between $\alpha$ and exaggeration remains for future work.} 
\label{fig-isolateddigits}
\end{figure*}

\end{document}